\documentclass[letterpaper]{article} 
\newif\ifExtendedVersion
\ExtendedVersiontrue 

\usepackage[]{aaai23}  
\ifExtendedVersion 
\nocopyright
\fi

\usepackage{times}  
\usepackage{helvet}  
\usepackage{courier}  
\usepackage[hyphens]{url}  
\usepackage{graphicx} 
\usepackage{natbib}  
\usepackage{caption} 
\usepackage{ifthen}
\newboolean{include_append}
\setboolean{include_append}{true} 

\usepackage[boxruled]{algorithm2e} 
\usepackage{natbib} 
\usepackage{amssymb} 
\usepackage{enumitem} 
\usepackage{booktabs} 
\usepackage{amsmath} 
\usepackage{color} 
\usepackage[dvipsnames]{xcolor} 
\usepackage{xspace}

\usepackage[size=tiny, textwidth=1.5cm]{todonotes}

\title{Rolling Horizon based Temporal Decomposition for the Offline Pickup and Delivery Problem with Time Windows}

\usepackage[normalem]{ulem}

\author{
    Youngseo Kim \textsuperscript{\rm 1}, Danushka Edirimanna \textsuperscript{\rm 1}, Michael Wilbur \textsuperscript{\rm 2}, Philip Pugliese \textsuperscript{\rm 3}, \\ Aron Laszka \textsuperscript{\rm 4}, Abhishek Dubey \textsuperscript{\rm 2}, Samitha Samaranayake \textsuperscript{\rm 1}
}
\affiliations{
    \textsuperscript{\rm 1}Cornell University, \textsuperscript{\rm 2}Vanderbilt University, \textsuperscript{\rm 3}Chattanooga Area Regional Transportation Authority,\\ \textsuperscript{\rm 4}Pennsylvania State University
}

\ifExtendedVersion
\usepackage{fancyhdr}
\fi

\begin{document}

\pagestyle{plain}

\ifExtendedVersion
\pagestyle{fancy}
\renewcommand{\headrulewidth}{0pt}
\lhead{Published in the proceedings of the 37th AAAI Conference on Artificial Intelligence (AAAI 2023).}
\rhead{}
\cfoot{\thepage}
\setlength{\headsep}{2.5em}
\setlength{\footskip}{3.5em}
\setlength{\headheight}{0em}
\fi

\setlength{\marginparwidth}{1.5cm}

\maketitle

\begin{abstract}
The offline pickup and delivery problem with time windows (PDPTW) is a classical combinatorial optimization problem in the transportation community, which has proven to be very challenging computationally. Due to the complexity of the problem, practical problem instances can be solved only via heuristics, which trade-off solution quality for computational tractability. Among the various heuristics, a common strategy is problem decomposition, that is, the reduction of a large-scale problem into a collection of smaller sub-problems, with spatial and temporal decompositions being two natural approaches. While spatial decomposition has been successful in certain settings, effective temporal decomposition has been challenging due to the difficulty of \textit{stitching} together the sub-problem solutions across the decomposition boundaries. In this work, we introduce a novel temporal decomposition scheme for solving a class of PDPTWs that have narrow time windows, for which it is able to provide both fast and high-quality solutions. We utilize techniques that have been popularized recently in the context of online dial-a-ride problems along with the general idea of rolling horizon optimization. To the best of our knowledge, this is the first attempt to solve offline PDPTWs using such an approach. To show the performance and scalability of our framework, we use the optimization of paratransit services as a motivating example. Due to the lack of benchmark solvers similar to ours (i.e., temporal decomposition with an online solver), we compare our results with an offline heuristic algorithm using Google OR-Tools. In smaller problem instances (with an average of 129 requests per instance), the baseline approach is as competitive as our framework. However, in larger problem instances (approximately 2,500 requests per instance), our framework is more scalable and can provide good solutions to problem instances of varying degrees of difficulty, while the baseline algorithm often fails to find a feasible solution within comparable compute times. 
\end{abstract}

\section{Introduction}

\begin{table*}[h]
\small
\caption{Recent literature for PDPTW variants in large scale network (published after 2018)}
\begin{tabular}{@{}p{3cm}llllllllllll@{}}
\toprule
                        & \begin{tabular}[c]{@{}l@{}}Depots\end{tabular} & \begin{tabular}[c]{@{}l@{}}Trips\end{tabular} & \begin{tabular}[c]{@{}l@{}}Vehicles\end{tabular} & Fleet & \begin{tabular}[c]{@{}l@{}}Vehicle \\ capacity\end{tabular} & \begin{tabular}[c]{@{}l@{}}Time \\ windows \end{tabular} & \begin{tabular}[c]{@{}l@{}}Ride \\ time\end{tabular} & \begin{tabular}[c]{@{}l@{}}Route \\ duration\end{tabular} & \begin{tabular}[c]{@{}l@{}}Selective \\ visits\end{tabular} & \begin{tabular}[c]{@{}l@{}}Obj.$^a$\end{tabular} & \begin{tabular}[c]{@{}l@{}}Static\\ /Dyn.$^b$\end{tabular} & \begin{tabular}[c]{@{}l@{}}\end{tabular} \\ \midrule
\cite{masmoudi2018dial}        & S                                                                         & S                                                                        & M                                                                           & HE    & \checkmark                                  & \checkmark                              &                                                      & \checkmark                                &                                                             & S                                                                             & Static  \\
\cite{sayarshad2018scalable}   & S                                                                         & M                                                                        & M                                                                           & HO    & \checkmark                                  &                                                         &                                                      &                                                           & \checkmark                                  & M                                                                             & Dyn.      \\
\cite{tellez2018fleet}         & M                                                                         & S                                                                        & M                                                                           & HE    & \checkmark                                  & \checkmark                              & \checkmark                           &                                                           &                                                             & S                                                                             & Static \\
\cite{luo2019two} & S                                                                         & M                                                                        & M                                                                           & HE    & \checkmark                                  & \checkmark                              & \checkmark                           &                                                           & \checkmark                                  & M                                                                             & Static    \\
\cite{bongiovanni2019electric} & M                                                                         & M                                                                        & M                                                                           & HE    & \checkmark                                  & \checkmark                              & \checkmark                           & \checkmark                                &                                                             & S                                                                             & Static  \\
\cite{liang2020automated}      & M                                                                         & S                                                                        & M                                                                           & HO    & \checkmark                                  & \checkmark                              & \checkmark                           & \checkmark                                & \checkmark                                  & S                                                                             & Dyn.   \\
\cite{malheiros2021hybrid}     & M                                                                         & S                                                                        & M                                                                           & HE    & \checkmark                                  & \checkmark                              & \checkmark                           & \checkmark                                &                                                             & S                                                                             & Static   \\
\cite{rist2021new}             & S                                                                         & S                                                                        & M                                                                           & HO    & \checkmark                                  & \checkmark                              & \checkmark                           & \checkmark                                &                                                             & S                                                                             & Static  \\
Ours & M  & M                                                                        & M                                                                           & HE    & \checkmark                                  & \checkmark                              & \checkmark                           & \checkmark &  \checkmark                                                           & M                                                                             & Dyn.  \\  \bottomrule
\end{tabular}
\label{tab:pdptwliterature}
\\
\footnotesize{$^a$ Abbreviation for objective, $^b$ Abbreviation for dynamic, S is an abbreviation for single, M is an abbreviation for multiple, HE is an abbreviation for heterogenous, HO is an abbreviation for homogenous.}\\
\end{table*}

The pickup and delivery problem with time windows (PDPTW) is a challenging optimization problem \cite{vrptw}. The PDPTW is a generalization of the vehicle routing problem with time windows (VRPTW) in which requests include both pickup and delivery time windows \cite{dumas1991pickup}. Furthermore, the PDPTW problem can be categorized as offline or online. In the offline setting, trip requests are gathered ahead of time and vehicle routes are optimized in advance for the day. In the online setting, requests are scheduled in real time as they arrive. In this work, we focus on the offline problem. Due to computational challenges, current state-of-the-art is focused on heuristic approaches, which often compromise solution quality for scalability \cite{vrptw}.
Previous research, motivated by real-world applications, has addressed various practical considerations, such as utilizing multiple depots and vehicles, enforcing strict time windows, allowing selective visits to customers, or optimizing multiple objectives at once \cite{ho2018survey, ropke2007models}. These approaches aim to address specific cases of the offline PDPTW and therefore often lack flexibility.

For solving the VRPTW, decomposition is a common strategy---dividing the original large-scale problem into a number of smaller sub-problems with respect to time or space. In spatial decomposition, requests are clustered by location. Then, decomposed problems are solved independently and solutions are merged together. There is significant literature on spatial decomposition based on the cluster-first and route-second principle \cite{ouyang2007design, desaulniers2002vrp}. In temporal decomposition, the time axis is split into multiple time intervals, and the original problem is strictly divided into small problems corresponding to these intervals. While some papers incorporate temporal information on top of a spatial decomposition \cite{bent2010spatial, qi2012spatiotemporal, tu2015novel}, we are not aware of successful approaches with pure temporal decomposition technique. Hence, we introduce a novel temporal decomposition. Na\"ive spatial or temporal decomposition can significantly degrade the solution as routes are isolated to individual spatial partitions or time bins. Also, it is non-trivial to stitch routes together that are independently obtained for each time interval in a post processing step \cite{zheng2019fuzzy}. Our temporal decomposition overcomes this issue by creating overlapping time bins and solving the problem via a rolling horizon approach. We note that a similar approach with spatial decomposition is much harder in pickup and delivery problems as each request is defined spatially in a four dimensional space (the product of two dimensional coordinates for the origin and destination nodes), while decomposing time only involves a single dimension. 

We consider a different approach for temporal decomposition based on rolling horizon optimization, where instead of dividing the problem into non-overlapping time intervals we iteratively solve the problem over a sequence of overlapping intervals. More precisely, we pick a time window size $T_w$ and a step size $t_s$, and create a sequence of sub-problems corresponding to time windows $\{(0, T_w), (t_s, t_s + T_w), (2t_s, 2t_s + T_w), \dots\}$. This approach eliminates the problem of boundary stitching and achieves smooth temporal transitions since part of the solution from one time interval can be updated in the next iteration. The time window ($T_w$) and step size ($t_s$) are hyperparameters that control the desired trade-off between computational efficiency and solution quality. This approach for temporal decomposition is also flexible enough to incorporate practical problem considerations, since adding complexity to a smaller problem instance (i.e., the sub-problem) is more computationally tractable than doing so with the full problem instance. 

The drawback of this approach is the need to solve a large number of sub-problems ($T/t_s$ instead of $T/T_w$) of size $T_w$. Thus, obtaining fast computation times requires a fast PDPTW solver. This leads us to consider recent approaches utilized in the online PDPTW literature and in particular an approach that works extremely well when the time windows are narrow~\cite{alonso2017demand}, which is a common characteristic in many passenger centric applications. To the best of our knowledge, this is the first attempt to deploy an online algorithm in conjunction with rolling horizon optimization for solving an offline PDPTW. We empirically show the performance and scalability of the rolling horizon framework through experiments. 

The remainder of this article is structured as follows. In Section \ref{section:literature}, we review variants of PDPTW and their practical considerations, and heuristics that have been developed for PDPTW. In Section \ref{section:method}, we explain the mathematical definition of the rolling horizon framework and the online solver that it is built upon. In Sections \ref{section:experiment}-\ref{section:results}, we show the performance and scalability of our framework through experiments on paratransit scheduling problems utilizing two sources of data. Finally, in Section \ref{section:conclusion}, we provide concluding remarks and discuss possible future directions.

\section{Literature Review}\label{section:literature}

\subsection{PDPTW variants for real-world application}\label{section:variants}

\citeauthor{ropke2007models} \shortcite{ropke2007models} provide a comprehensive survey of PDPTW solvers developed up to 2007. \citeauthor{ho2018survey} \shortcite{ho2018survey} conducted a comprehensive literature review for PDPTW variants published from 2007 to 2018 (see Table 5-8 in their paper \cite{ho2018survey}) and we provide an extended table for papers published after 2018 in Table \ref{tab:pdptwliterature}. Columns are all the typical features that have been considered in PDPTW variants, and a more detailed explanation of the features is explained in online appendix \cite{onlineappendix}. 

\citeauthor{ho2018survey} \shortcite{ho2018survey} pointed out the research gaps and encouraged the development of techniques that can be adapted to solve the many variants of the PDPTW. Our approach can or can be easily modified to consider all the typical features of the PDPTW variants shown in Table~\ref{tab:pdptwliterature}. For example, even though it is outside the scope of this work, our approach can be easily adapted to handle dynamic updates of an existing solution in real time. While reviewing papers published after 2018, we could not find any approaches that can efficiently accommodate as many practical considerations as our approach.

The flexibility of our framework stems from two types of decompositions. First, the online VRP algorithm ~\cite{alonso2017demand} that our approach is built upon decomposes the high-capacity vehicle-passenger matching problem into a routing problem and an assignment problem, and develops heuristics for effectively computing feasible routes when time windows are relatively tight. Second, the temporal decomposition allows for dividing the original problem into a sequence of more tractable sub-problems. The computational gains achieved via these decompositions allow for solving more complex problem variants.

\subsection{Solution methods for PDPTW}

Previous literature has proposed different solution approaches to solve PDPTW and its variants. The vast majority of exact algorithms are developed using techniques such as branch-and-cut \cite{cordeau2006branch, ropke2007models}, branch-and-price \cite{garaix2010vehicle}, and branch-and-price-and-cut \cite{qu2015branch}. Exact methods are unable to solve large instances within a reasonable time due to the intrinsic hardness of PDPTWs. \cite{ropke2009branch} obtained the exact solution up to 500 customers in some benchmark instances, but many of the benchmark instances remain unsolved in optimality \cite{ho2018survey}. Furthermore, the largest solvable instance size gets smaller for more complex variants of the problem. 

Spatial and temporal decomposition is one common strategy to tackle large scale problems. \citeauthor{bent2010spatial} \shortcite{bent2010spatial} suggested an iterative and adaptive decomposition scheme that defines sub-problems based on the spatiotemporal features of the existing solution, solves them independently, and inserts them into the existing solution. \citeauthor{qi2012spatiotemporal} \shortcite{qi2012spatiotemporal} decided sub-problems using a clustering method based on spatiotemporal distance, and merged them after solving independently. \citeauthor{tu2015novel} \shortcite{tu2015novel} defined a spatial-temporal distance and used it to speed up a local search heuristic. However, all of these techniques explicitly depend on the problem being a VRPTW, and cannot be applied to situations involving pickup and delivery. 

While there are no attempts to introduce temporal decomposition in PDPTW, various heuristics and metaheuristics have been developed for solving real-world scale instances. These methods include insertion based heuristics~\cite{luo2007rejected,hame2011adaptive}, tabu search \cite{kirchler2013granular, detti2017multi}, simulated annealing \cite{braekers2014exact}, variable neighborhood search \cite{parragh2010variable}, large neighborhood search \cite{ropke2006adaptive}, and genetic algorithms \cite{jorgensen2007solving}. In general, it is hard to compare performance among these algorithms due to the lack of uniformity in the specific problem variants they consider and the use of different benchmark instances of different sizes. 

\begin{figure*}[h!]
  \centering
  \includegraphics[width=\textwidth]{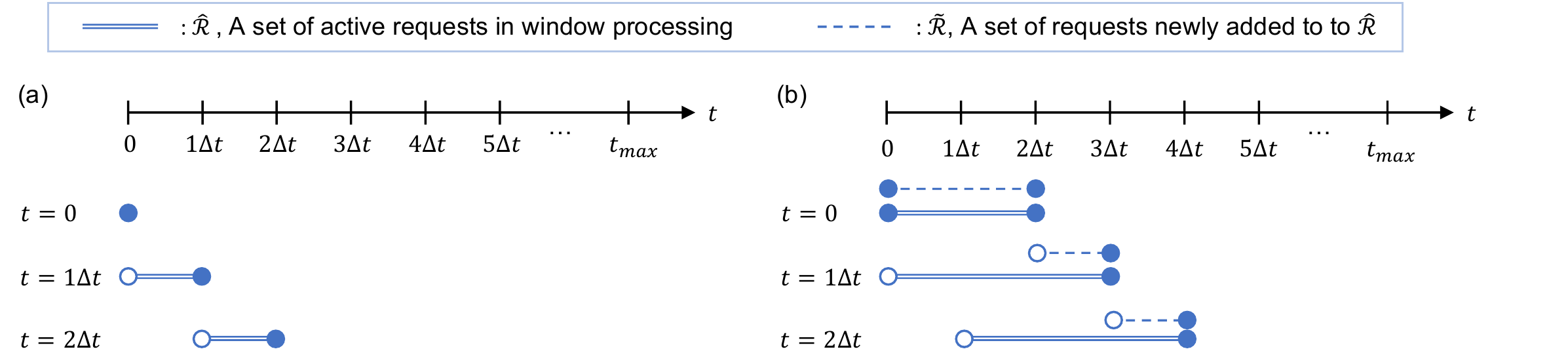}
  \caption{Window processing in (a) online (b) rolling horizon framework (when sliding window size is $3\Delta t$ and step size is $\Delta t$)}
  \label{fig:RH}
\end{figure*}

\section{Methodology}\label{section:method}

\newcommand{\DesirePickupTime}[1]{\ensuremath{\hat{t}^{\textit{pickup}}_{#1}}}
\newcommand{\EarliestDropoffTime}[1]{\ensuremath{\hat{t}^{\textit{dropoff}}_{#1}}}
\newcommand{\ActualPickupTime}[1]{\ensuremath{t^{\textit{pickup}}_{#1}}}
\newcommand{\ActualDropoffTime}[1]{\ensuremath{t^{\textit{dropoff}}_{#1}}}

\newcommand{\PromisedPickupTime}[1]{\ensuremath{\bar{t}^{\textit{pickup}}_{#1}}}
\newcommand{\PromisedDropoffTime}[1]{\ensuremath{\bar{t}^{\textit{dropoff}}_{#1}}}

\newcommand{\TripSet}[0]{\ensuremath{{\cal{T}}}\xspace}
\newcommand{\RequestSet}[0]{\ensuremath{{\cal{R}}}\xspace}
\newcommand{\ActiveRequestSet}[0]{\ensuremath{\hat{\cal{R}}}\xspace}
\newcommand{\NewRequestSet}[0]{\ensuremath{\tilde{\cal{R}}}\xspace}
\newcommand{\VehicleRoutes}[0]{\ensuremath{\mathbb{R}}\xspace}

\newcommand{\VehicleSet}[0]{\ensuremath{{\cal{V}}}\xspace}
\newcommand{\PassengerList}[0]{\ensuremath{{\cal{P}}_{v}}\xspace}
\newcommand{\PassengerForVehicle}[1]{\ensuremath{{\cal{P}}_{v_{#1}}}\xspace}
\newcommand{\Assignment}[0]{\ensuremath{\textit{Assignments}}\xspace}

\newcommand{\EdgeTripVehicle}[0]{\ensuremath{\epsilon_{ij}}\xspace}
\newcommand{\EdgeRequestTrip}[0]{\ensuremath{{\chi_k}}\xspace}
\newcommand{\EdgeTripVehicleSet}[0]{\ensuremath{{\cal{E}}_{TV}}\xspace}

\newcommand{\MaxWaiting}[0]{\ensuremath{W_{\textit{max}}}\xspace}
\newcommand{\MaxDelay}[0]{\ensuremath{D_{\textit{max}}}\xspace}
\newcommand{\RHFactor}[0]{\ensuremath{c^{RH}}\xspace}

\subsection{Problem input and output}\label{sec:probleminput}
We assume a set of vehicles $\VehicleSet = \{v_1, \ldots, v_m\}$ with a fixed vehicle-specific capacity and a set of requests $\RequestSet = \{r_1, \ldots, r_n\}$, where each request $r_k$ contains a pickup and drop-off locations, and desired pickup time \DesirePickupTime{k}. We calculate \EarliestDropoffTime{k}, the earliest possible drop-off time, by adding the estimated travel time between pickup and drop-off locations to the desired pickup time. Each problem instance is also defined by the following set of parameters: maximum allowed delay time \MaxDelay and waiting time \MaxWaiting of customers. The time horizon of the problem $t_{max}$ is divided into smaller intervals with length $T_w$, which we refer to as the sliding window size. The rolling horizon optimization also defines a step size $t_s$, which in our approach is equal to the \textit{batch size} of the online optimization algorithm (see below). Correspondingly, we pick a sliding window size that is a multiple of the batch size, which we can interpret as a look-ahead window from the point of view of the online optimization problem. Selecting an appropriate step size and sliding window size is part of our investigation. The output of the model is a set of vehicle routes $\VehicleRoutes = \{R_1, R_2, \ldots, R_m\}$, where each route is an ordered set of pickup and drop-off locations and estimated time to visit. We obtain vehicle routes that maximize service rate while minimizing vehicle miles traveled as a secondary objective.

\subsection{Computational approach}

In the rolling horizon framework, sub-problems are solved in sequential order along the time axis, with the solution to each sub-problem determining the best passenger-vehicle matching for the corresponding time interval. The solution of the first $t_s$ minutes is fixed, but the remainder ($T_w - t_s$) can be reoptimized in the next iteration. We use the term batch to refer to the set of requests belonging to a given sub-problem, i.e, the requests whose desired pickup time is within the sliding window corresponding to that sub-problem. The process for selecting the requests corresponding to each batch is called window processing, which is illustrated in Figure \ref{fig:RH}. The online approach with $T_w = t_s$ is a myopic strategy because each batch only includes requests that have already entered the system. In contrast, our rolling horizon framework can consider batches that extended into the future (known exactly in our case because this is an offline problem and all the demands are known). After solving the optimization problem for a given sub-problem, the routing schedule is finalized for all the requests that will not appear in the next sub-problem. In the next sub-problem, we solve another optimization problem with a new time interval which is shifted forward one step (i.e., by $t_s$).

\subsection{Problem formulation and algorithm}

The entire algorithm follows an iterative process. In each sub-problem, an optimization problem considers an active request set \ActiveRequestSet ($\subset$\RequestSet). New requests obtained from the window processing are added to the active requests set. Requests that have been picked up by any vehicle are removed from the active request set at the end of each iteration. To track requests that have already been picked up, we maintain the set ${\cal{P}}_{v,t}$ which is the set of passengers that have boarded onto vehicle $v \in \VehicleSet$ prior to time $t$. Then, we solve the matching problem (RTV-ILP) between the active request set \ActiveRequestSet and the vehicle set \VehicleSet. As a result, we obtain the order and the scheduled time to pickup and drop-off customers for each vehicle. Then, the discrete event vehicle simulator receives the vehicle location from the prior sub-problem, executes scheduled routes from the optimizer until the current simulation time, updates travel times, and then finalizes the vehicle locations for the next sub-problem. From the vehicle simulation, we get a set of vehicle routes until the current time. The finalized routes in the current iteration cannot be changed in the later iterations. We describe the process in more detail in the following sections. Pseudo code for the entire algorithm and window processing is provided in the online appendix \cite{onlineappendix}. 

\subsubsection{Window processing.} The procedure for selecting a set of requests to be considered based on their desired pickup time, \NewRequestSet. As discussed earlier, the sliding window size $T_w$ and step size $t_s$ are part of the investigation. In general, increased sliding window size leads the better solution quality but exponentially increases computational time. Step size $t_s$ can be the overlapped window size ($T_w - t_s$), which will also increase solution quality. Decreasing the step size linearly increases computational time as we need to solve around $T/t_s$ sub-problems.

\subsubsection{RTV-ILP.} The integer linear program (ILP) for assigning requests to trips and trips to vehicles. The RTV-ILP framework can solve fairly large problem instances, which accommodates larger sub-problems in the temporal decomposition. The illustration of the RT-V structure can be found in the online appendix \cite{onlineappendix}. We refer readers to~\cite{alonso2017demand} for more detail. Requests \RequestSet are aggregated into trips \TripSet based on service constraints. The RT-V graph contains all feasible trip-vehicle pairings, \EdgeTripVehicleSet. The existence of an edge between a trip $T_i \in$ \TripSet and a vehicle $v_j \in$  \VehicleSet indicates that it is feasible for vehicle $j$ to serve trip $i$. The feasibility of a trip is determined by whether all the requests that belong to the trip can be served by a vehicle while satisfying the constraints in Equations \ref{eqn:max_waiting} and \ref{eqn:max_delay}. The constraints ensure that i) the waiting time is not greater than the maximum waiting time \MaxWaiting, and ii) the delay time is not greater than the maximum delay time \MaxDelay. Recall that \ActualPickupTime{k} denotes the actual pickup time and \DesirePickupTime{k} denotes the desired pickup time. Also, \ActualDropoffTime{k} denotes the actual dropoff time and \EarliestDropoffTime{k} denotes the earliest possible dropoff time.

\begin{flalign}
    \ActualPickupTime{k}-\DesirePickupTime{k} & \leq \MaxWaiting, \forall k \label{eqn:max_waiting} \\
    \ActualDropoffTime{k}-\EarliestDropoffTime{k} & \leq \MaxDelay, \forall k \label{eqn:max_delay}
\end{flalign}

\noindent Note that in the case of high-capacity vehicles, enumerating all possible combinations of pickups and deliveries is still computationally expensive. Therefore, we use an insertion-based heuristic when dealing with more than 4 passengers \cite{alonso2017demand, wilbur2022online}. This process can be further improved with heuristics that appropriately trade-off speed and accuracy.

After building the RT-V graph, we need to solve an ILP to obtain the optimal matching. 

\begin{align}
    & \operatorname*{argmin}_{\EdgeTripVehicle, \EdgeRequestTrip} 
    & & \sum_{ \{ij: \EdgeTripVehicle \in \EdgeTripVehicleSet\} }{c_{ij} \EdgeTripVehicle} + \sum_{k \in {\cal{R}}}{c_{k}} \EdgeRequestTrip &\\
    & \text{s.t.} & &  \sum_{ \{i: T_i \in \TripSet \}} \EdgeTripVehicle \leq 1 &, \forall v_j \in \VehicleSet \label{eqn:ilp_const1}\\
    & & & \sum_{ \{i: T_i \in \TripSet \}}\sum_{\{j: V_j \in \VehicleSet \} } \EdgeTripVehicle + \EdgeRequestTrip = 1 &, \forall r_k \in \RequestSet \label{eqn:ilp_const2}
\end{align}

\noindent Decision variables \EdgeTripVehicle and \EdgeRequestTrip are binary. If vehicle $j$ is assigned to trip $T_i$, $\EdgeTripVehicle = 1$; otherwise, $\EdgeTripVehicle = 0$. If the request $r_k$ cannot be served by any vehicle, $\EdgeRequestTrip = 1$; otherwise, $\EdgeRequestTrip = 0$. In the objective function, there is cost $c_{ij}$ which is the total vehicle miles traveled by vehicle $v_j$ when serving trip $T_i$. The other cost $c_k$ is a large constant to penalize unserved requests, such that the service rate is maximized. Thus, the objective function prioritizes maximizing the service rate and then minimizing the vehicle miles traveled as a secondary objective. Equation \ref{eqn:ilp_const1} guarantees that a vehicle is assigned to at most one trip. Equation~\ref{eqn:ilp_const2} imposes that each request should be either served by a vehicle or ignored.

An RT-V graph is built at each sub-problem and the best passenger-vehicle matching can be updated until the passenger is picked up or their maximum waiting time is exceeded. In other words, passengers can be swapped to another vehicle prior to pickup. However, additional constraints enforce that a previously matched passenger cannot be ignored. 

\section{Experimental Design}\label{section:experiment}

We showcase the performance of our framework using paratransit scheduling as a motivating application. Paratransit is a service for passengers who are unable to use fixed-route transit, which is provided by public transit agencies as mandated by the Americans with Disabilities Act. Due to the large costs associated with operating paratransit services, transit agencies are constantly looking to improve service efficiency and provide a high level of service at a lower cost. Customers book paratransit trips either by phone or via an app, and can make a reservation as early as two weeks in advance or within a few days of travel. Customers are given a confirmation of travel with an estimated pickup and drop-off time shortly after the booking request. The service is required to provide a tight pickup time and drop-off window, typically around 30 minutes. Additionally, paratransit consists of high capacity vehicles which adds to the computational complexity of the problem. Some paratransit services allow customers to request trips in real time, but service is not guaranteed for these trips. Accordingly, the paratransit problem is an offline optimization problem because most service providers require reservations to be made by the end of the previous day, with a few exceptions allowing reservations to be placed during the service day.

We obtain data from two different sources for experiments. First, we use paratransit trip requests provided by a public transit agency. Second, we use New York City taxi \cite{nyctaxidata} data to investigate the scalability of our approach with a larger dataset. The parameters that we use for the experiments can be found in the online appendix \cite{onlineappendix}. We provide two baselines. The first baseline is a fully online solver while the second baseline is an offline heuristic solver.

\subsubsection{Real-world paratransit data}

We used 6 months of paratransit trip requests between January 1, 2021 and June 30, 2021. The data is provided by our partner agency, the \textit{Chattanooga Area Regional Transportation Authority (CARTA)}, which is a mid-sized public transit agency in the United States. As this dataset is directly from a transit agency, we use it to show the performance of our approach on real-world data. Each trip request contains pickup and dropoff locations and the requested pickup time. For privacy considerations, the location information is anonymized as follows. The service area is discretized into a grid of one square mile tiles. For each request, the service location is shifted to a random location within the corresponding cell. Thus, all the demands from a given cell are randomly redistributed within that cell. The requests are also temporally aggregated by the agency to 15 minutes. Thus, we set the step size to 15 minutes, which is the step size that we can use. We randomly selected 30 weekdays for the experiments. There were an average of 172 trip requests per instance in this dataset.

\subsubsection{New York City taxi data}

Recall that a primary advantage of our approach is its ability to scale to a large number of requests. To test this, we acquired 31 days of taxi trip requests (January 1, 2016 to January 30, 2016) for New York City \cite{nyctaxidata}. To better represent paratransit trips, we deleted short-distance trips which are less than 5 km. We randomly sampled 1\% and 20\% of the NYC taxi data to generate two new paratransit datasets for our investigation. We sampled 1\% of the data to have a similar size of data to the real-world paratransit dataset, having an average of 129 trips per instance. The 20\% dataset had an average of 2,587 requested trips per instance and was used to investigate the scalability of our approach. We refer to the 1\% sampled data as Scenario 1 and the 20\% sampled data as Scenario 2. A detailed description of the data statistic is provided in \ifExtendedVersion{Appendix \ref{section:datadescription}}\else{the online appendix \cite{onlineappendix}}\fi. As New York City taxi data is larger than the paratransit data, we set the step size to 5 minutes.

\subsection{Metrics}

For each day, we calculate three metrics to evaluate the performance of our approach compared to the baselines. Service rate is the number of requests served divided by the total number of requests in a day. The compute time per request is the total time the solver takes to run for a day divided by the number of requests on that day. The average delay time is the average time difference between the actual dropoff time and the earliest possible dropoff time that is calculated by adding the shortest travel time to the requested pickup time. 

\subsection{Baselines}\label{section:baseline}
Our approach provides an efficient, tune-able approach that provides a better trade-off between computational efficiency and solution quality. To investigate this trade-off, we implement a fully online, myopic solution and a fully offline heuristic solver. 

\subsubsection{Online Solver}

We use our own approach with a sliding window size $T_w$ to be equal to the step size $t_s$ to represent the online solver. By having an overlapped window size ($T_w - t_s$) of zero, our solver is simplified to a purely online approach as it does not utilize future knowledge. We refer to this baseline as RH0 comparing with our approach RH1, RH2, RH3, which have overlapped windows whose size is a multiple of step size of 1, 2, and 3, respectively.

\subsubsection{Offline Solver}

We implemented an offline PDPTW benchmark solver in Google OR-Tools, a well established, modern, and publicly available VRP solver developed by Google \cite{ortools}. We use guided local search (GLS) which is known as the best performing setting for the OR-Tools PDPTW solver as the baseline heuristic. The general GLS approach was first proposed by \citeauthor{kilby1999guided} \shortcite{kilby1999guided} as an efficient anytime heuristic, that aims to iteratively improve the solution for a fixed set of time. GLS requires an initial solution to improve upon. To find the initial solution we used a parallelized cheapest insertion approach which iteratively builds a solution by inserting the cheapest node at its cheapest position without violating the PDPTW constraints; cost is defined by the objective function \cite{ortools}. We optimize for service rate by setting a sufficiently large penalty for trips that are not serviced. The secondary objective is to minimize passenger travel time. The baseline includes constraints on pickups and dropoffs, as well as time windows (as in our approach). 

In addition, in the online appendix \cite{onlineappendix} we also compare our algorithm with the LKH-3 solver that uses a modified Lin-Kernighan-Helsgaun heuristic. This is a state-of-the-art solver for constrained traveling salesman and vehicle routing problems.

\subsection{Reproducibility}
All the codes and datasets we used for experiments are available online. Software code for the rolling horizon framework is available in the following repository: \url{https://github.com/MAS-Research/RollingHorizon.git}. Code for the baseline offline heuristic using Google OR-Tools is available in the following: \url{https://github.com/MAS-Research/RollingHorizon_baseline_ORTools}.

\section{Results}\label{section:results}

\subsection{Evaluation on real-world paratransit dataset}

\begin{figure}[h!]
  \centering
  \includegraphics[scale=0.31]{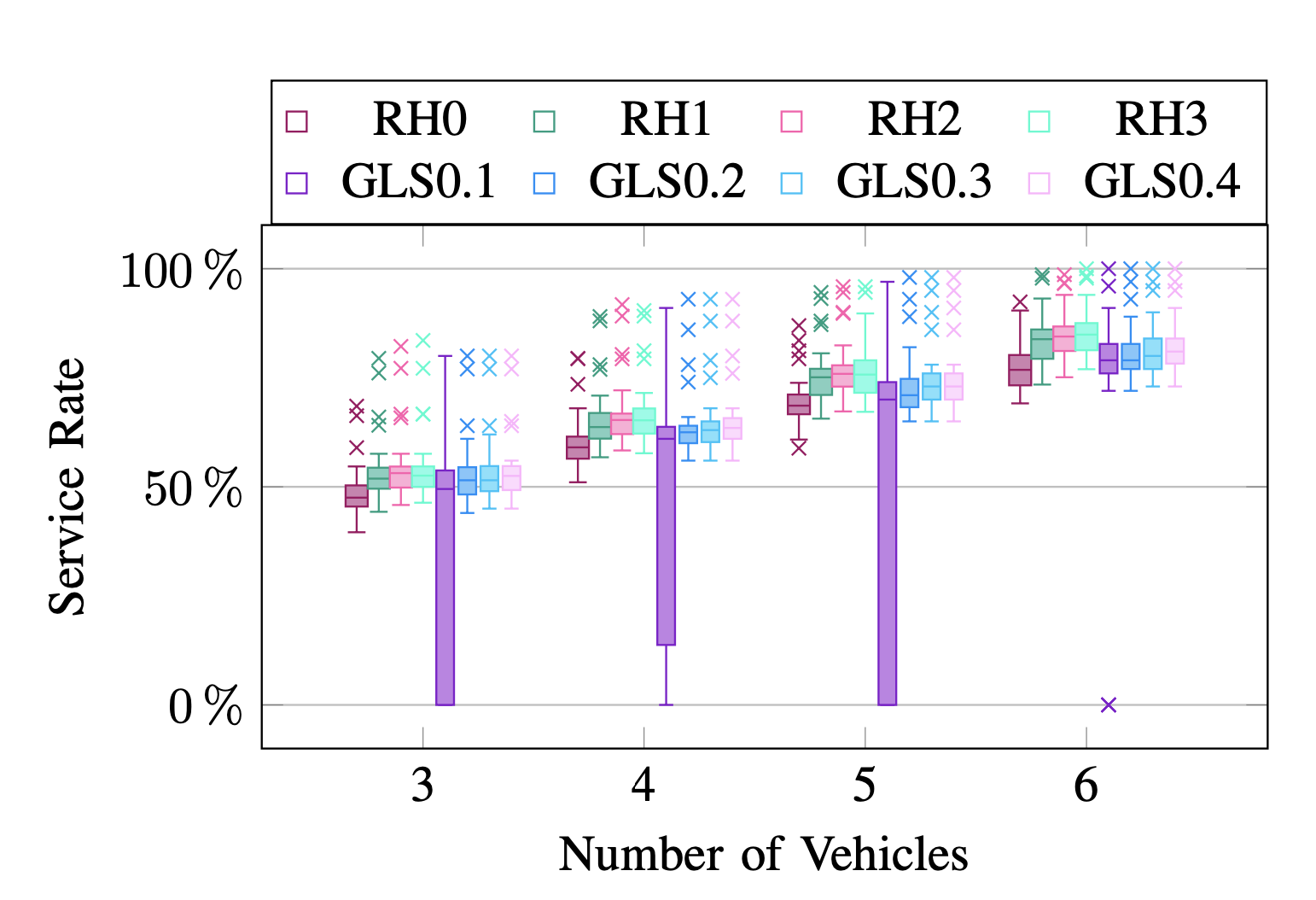}
  \caption{Service rate, Chattanooga paratransit dataset}
  \label{fig:ChattaServiceRate}
\end{figure}

Figure \ref{fig:ChattaServiceRate} provides service rates for the 31 day experiment. Overall, our framework (RH1, RH2, RH3) outperforms the online approach (RH0). In an experiment with 6 fleets, RH3 brings an increase in service rate from 77.7\% to 85.6\% compared to RH0, which corresponds to a 10.1\% increase. We observe that the improvement of the service rate reduces as the sliding window size increases. This is because future information has a more significant impact on current decisions when the events are imminent. However, computational time exponentially increases as we increase the sliding window size (The result can be found in the online appendix \cite{onlineappendix}.). In practice, operators can decide the appropriate sliding window size and step size to trade off service rate and compute time. 
 
\subsection{Evaluation of NYC datasets}

\begin{figure*}[h!]
  \centering
  \includegraphics[width=0.98\textwidth]{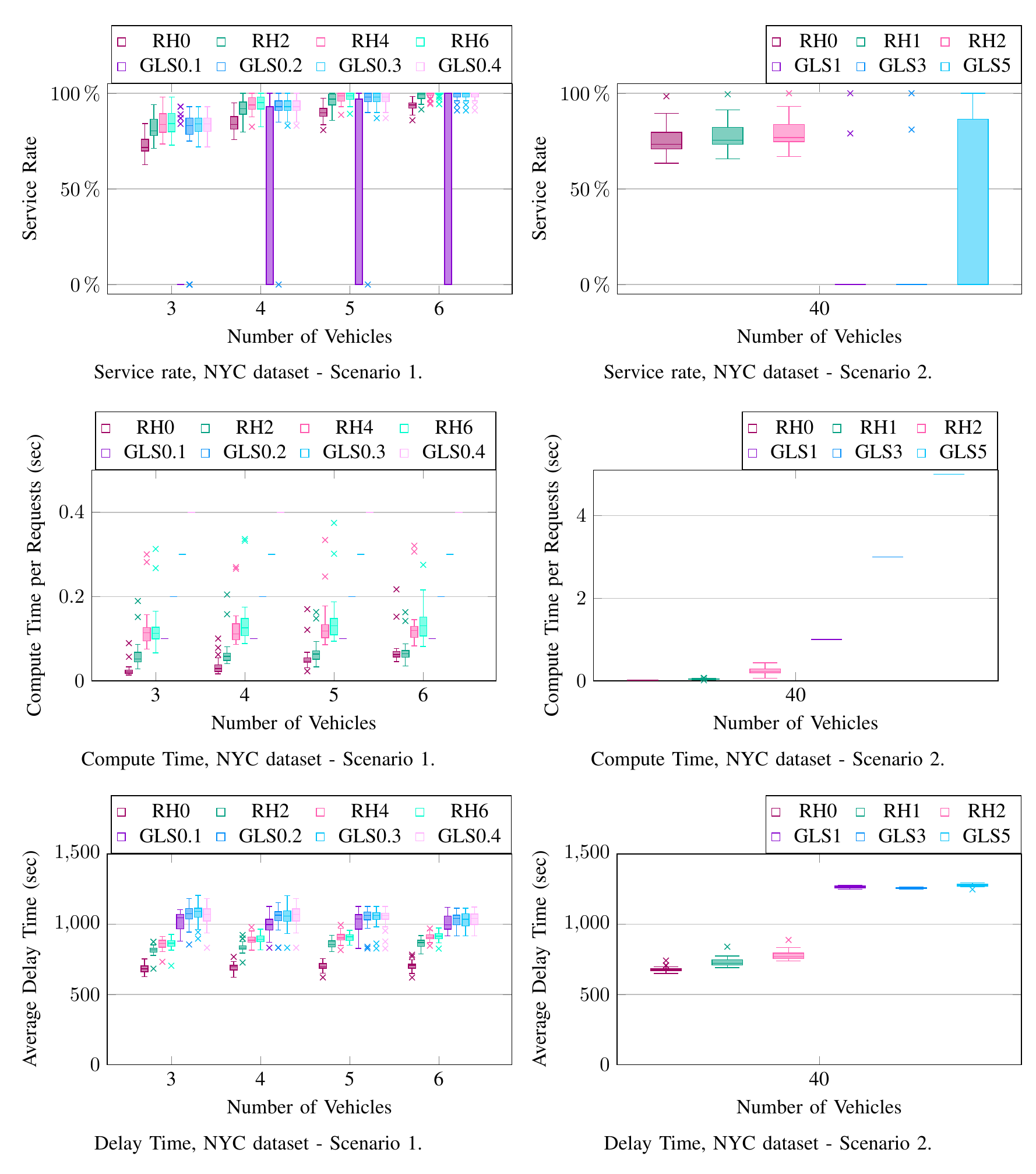}
  \caption{Comparison of the rolling horizon framework with the baseline approach. RH$X$ denotes the rolling horizon solutions with  $X$ indicating the rolling horizon factor. GLS$Y$ denotes the guided local search solutions with the $Y$ indicating time limit.}
  \label{fig:NYC}
\end{figure*}

In Figure \ref{fig:NYC}, the left column shows the results of scenario 1 (with an average of 129 requests per instance) and the right column shows the results of scenario 2 (with an average of 2,587 requests per instance). In scenario 1, the performance of the rolling horizon framework is as good as that of GLS. Recall that GLS is an anytime algorithm from which we obtain the best among the solutions that have been found within a time limit. We obtain the best solution of GLS after the given time limit of 0.1, 0.2, 0.3, and 0.4 seconds per request. GLS fails to find a feasible solution in some instances with 0.1, 0.2 seconds time limit, and starts to provide results to all instances with 0.3 seconds time limit. More specifically, with a 0.1 second time limit, GLS cannot get a feasible solution in 25, 22, 21, 20 instances among 31 instances with 3, 4, 5, 6 vehicles, respectively. With a 0.2 second time limit, GLS cannot get a feasible solution in 4, 1, 1 instances among 31 instances with 3, 4, 5 vehicles, respectively. 

In our framework, the mean of service rates increases as the sliding window size increases (RH0, RH2, RH4, RH6, in that order), and it has a saturation point in all experiment settings with varying fleet sizes. The rolling horizon approach achieves saturation points when the time limit becomes 0.3 seconds per request, which is the time limit that GLS starts performing well. The saturated service rates are close to or even exceed the service rates of GLS. This indicates that the rolling horizon framework obtains good enough solution quality which is comparable to that of GLS within similar compute times. Also, the average delay time from the rolling horizon framework is smaller than that of GLS, which would lead to better user experiences. 

Scenario 2 highlights the scalability of our framework. Recall that the number of requests in scenario 2 is 20 times larger than that of scenario 1. Our framework achieves reasonable service levels and average delay times. Compare to RH0 which shows 75.6\% of the average service rate and 98.4\% of the maximum service rate, RH2 shows improvement with 79.0\% of the average service rate and 100\% of the maximum service rate. Even RH2 whose compute times are the largest in our framework can solve any instance within 1 second per request. On the contrary, GLS cannot get any feasible solution in 29 among 31 instances within the time limit of 1 and 3 seconds per request. In a time limit of 5 seconds per request, GLS performs better but still cannot get any feasible solution in 16 instances. The time limit of 5 seconds per request corresponds to around 3.5 hours time limit for the entire instance, which is already quite long considering that operators need to obtain a schedule for the next day within a couple of hours.

\section{Conclusions}\label{section:conclusion}

The pickup and delivery problem with time windows (PDPTW) is a challenging operational problem, and several generalizations based on practical considerations make the problem even more complicated. In this paper, we introduce a new temporal decomposition scheme to solve the PDPTW at scale. Our approach uses a state-of-the-art online algorithm within a rolling horizon framework to solve a sequence of smaller sub-problems that collectively cover the entire original problem. This strategy avoids the primary pitfall of more na\"ive temporal decompositions, namely the challenge of stitching together sub-problems. The computational gains made through this decomposition provide us the flexibility to add additional features (corresponding to practical needs) even though this introduces extra complexity. 

We choose the paratransit scheduling problem to showcase the performance of our rolling horizon framework in different scales of networks with different demand profiles. In the real-world size instances, the rolling horizon framework achieves as high service rates as the benchmark offline solver within comparable computational times. Moreover, our framework can be scaled up to around 2,500 requests while the benchmark solver often fails to find any feasible solution within the 5 seconds time limit per request. Due to its computational efficiency, our approach can be used to continuously add new requests to the system after a service plan is made, giving paratransit operators more flexibility in adding last minute requests. Our framework can be also used to obtain a reasonably good initial feasible solution quickly, making it a good candidate to be combined with other local search heuristics that can subsequently improve the solution as time permits.

\section*{Acknowledgment}
This research is partially sponsored by the National Science Foundation under grants CNS-1952011 and CIS-2144127, and the Department of Energy under Award Numbers DE-EE0009212 and DE-EE0008464. We thank Matthew Zalesak for his assistance with implementing the proposed framework.

\bibliography{aaai23}

\ifExtendedVersion{

\clearpage
\section{Appendix}

\subsection{A list of practical considerations}\label{section:columns}

\begin{enumerate}[noitemsep,leftmargin=*] 
\item{Multi vehicle} utilizing multiple fleets.
\item{Multi depots} multiple starting and ending locations for the vehicle fleet.
\item{Multi trips} allowing a vehicle to return to a depot multiple times in a single day.
\item{Heterogeneous fleet} various combinations of equipment for different types of passengers.
\item{Vehicle capacity} limiting the maximum number of passengers. 
\item{Time window} user-specified earliest and latest time bound for pick-up and drop-off.\footnote{Although our framework does not explicitly have the concept of the time window, inputs of our model can translate to time windows. Desired pickup time/earliest drop-off time corresponds to the start of the time window. Then, the end of the time window can be set by adding the maximum waiting time/detour time. The definition of the inputs can be found in Section~\ref{sec:probleminput}.} 
\item{Ride time} limiting the maximum time difference between the scheduled pickup and dropoff time.
\item{Route duration} limiting the maximum time difference between the times of leaving from and returning to a depot.
\item{Selective service} allowing selective pickups and deciding which requests to accommodate.
\item{Multiple objectives} objective functions consisting of multiple measures. 
\item{Dynamic} allowing dynamic modification of existing plans in response to new information. 
\end{enumerate}

\subsection{RT-V graph}\label{section:rtv}

\begin{figure}[h]
  \centering
  \includegraphics[scale=0.5]{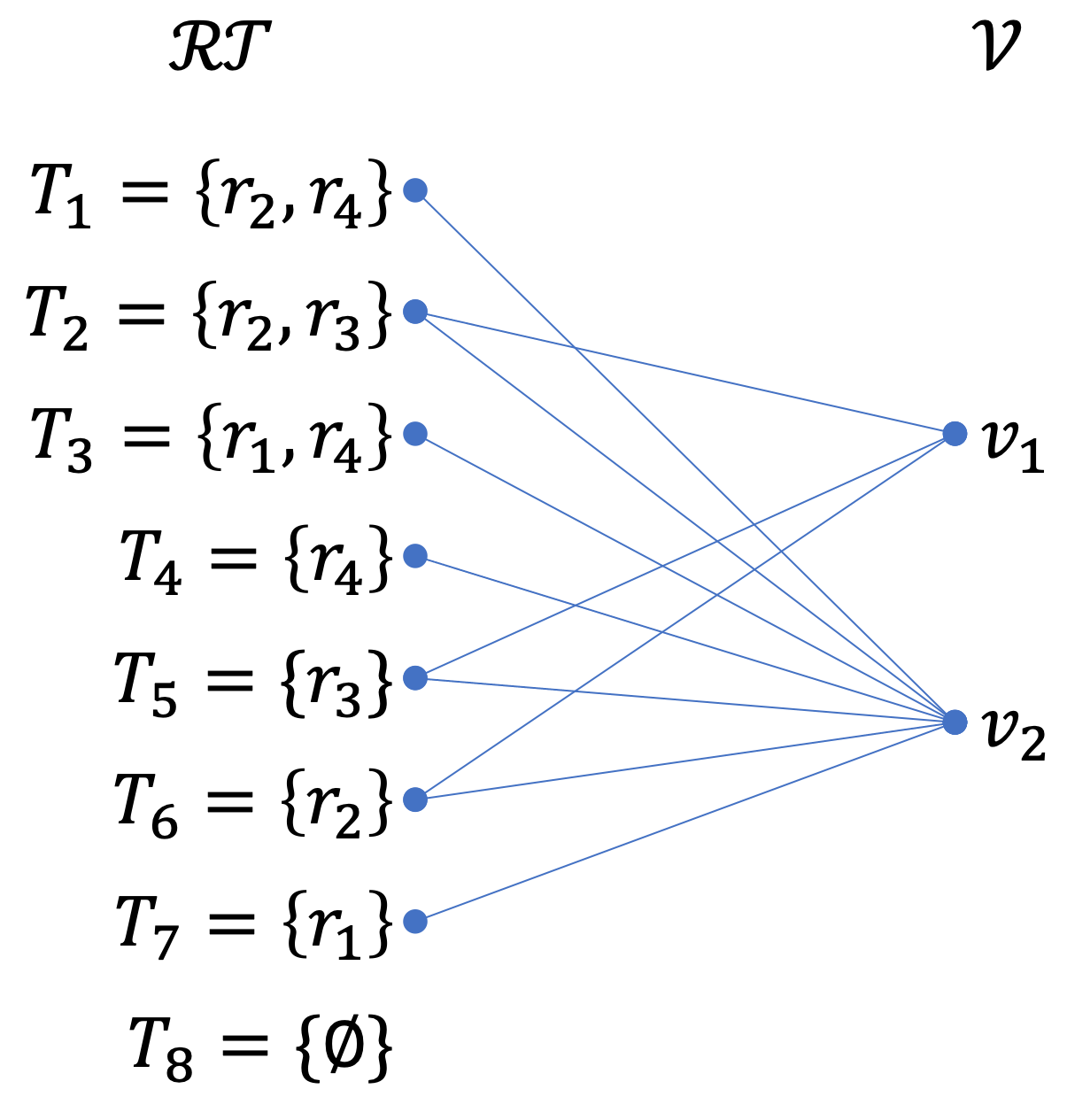}
  \caption{RT-V graph for an example instance (reconstructed from the schematic figure in \citeauthor{alonso2017demand}'s work)}
  \label{fig:RTVgraph}
\end{figure}

Figure \ref{fig:RTVgraph} illustrates the structure of RT-V graph. Requests $r_k \in$ \RequestSet are aggregated into trips \TripSet if it is feasible to serve those requests by one vehicle, in other words, if those requests are shareable. Edges \EdgeTripVehicleSet between a trip $T_i \in$ \TripSet and a vehicle $v_j \in$ \VehicleSet represent all possible candidate pairs between trip and vehicle meaning that it is feasible for the vehicle to serve all requests in the trip $T_i$.

\subsection{Data statistics}\label{section:datadescription}

An overview of the two datasets is provided in Table \ref{tab:statistics2}. The real-world paratransit dataset contains 30 instances. Each instance represents different operation days. There are an average of 172 requests per instance with a standard deviation of 33. NYC taxi data contains 31 instances. Scenario 1 contains 129 requests per instance with a standard deviation of 29. Scenario 2 contains 2587 requests per instance with a standard deviation of 570. 

\begin{table}[h]
\footnotesize
\centering
\caption{Overview of Datasets}
\begin{tabular}{|l|c|c|c|}
\hline
                                                                 & \bf \begin{tabular}[c]{@{}c@{}} Real-world \\[-0.2em] Dataset\end{tabular} & \bf \begin{tabular}[c]{@{}c@{}}NYC \\[-0.2em] Scenario 1\end{tabular} & \bf \begin{tabular}[c]{@{}c@{}}NYC \\[-0.2em] Scenario 2\end{tabular} \\ \hline
Number of Days                                                   & 30    & 31                                                        & 31                                                        \\ \hline
\begin{tabular}[c]{@{}l@{}}Mean Request \\ per instance\end{tabular}  & 172   & 129                                                       & 2587                                                      \\ \hline
\begin{tabular}[c]{@{}l@{}}Std. Dev. Requests \\ per instance \end{tabular} & 33    & 29                                                        & 570                                                       \\ \hline
\end{tabular}
\label{tab:statistics2}
\end{table}

\subsection{System parameters}\label{section:parameters}

We identified parameter settings based on discussions with the public transit agency, which are shown in Table \ref{tab:parameters}. The number of vehicles varies from 3 to 7 and each vehicle can accommodate up to 8 passengers at a time. Maximum waiting time is defined as the time difference between the actual pickup time and the desired pickup time which is set to 30 minutes. The maximum delay time is defined as the time difference between the actual dropoff time and the earliest possible dropoff time and is also set to 30 minutes. In practice, our partner agency set the same goal for the service levels. On average, our partner agency plans for 5-10 minutes to load/unload a passenger, thus we conservatively set the dwell time to 10 minutes. We set the step size to 15 minutes, which is the minimum interval that we can use because time information in the original data was aggregated by 15 minutes.

\begin{table}[h]
 \footnotesize
\centering
\caption{Parameter settings for real-world paratransit dataset}
\begin{tabular}{|l|l|}
\hline
\textbf{Parameter} & \textbf{Values} \\ \hline 
Fleet size & 3, 4, 5, 6, 7 \\ \hline
Vehicle capacity & 8 \\ \hline
Maximum waiting time & 30 (min) \\ \hline
Maximum delay time & 30 (min) \\ \hline
Dwell time & 10 (min) \\ \hline
Step size & 15 (min) \\ \hline
\end{tabular}%
\label{tab:parameters}
\end{table}

The parameter settings for the NYC taxi investigation are provided in Table \ref{tab:parameters_NYC}. For scenario 1 we vary the fleet size from 3 to 6 vehicles while for scenario 2 we investigate fleet sizes of 30, 40, 50, and 60. The vehicle capacity is again set to 8. Due to the higher frequency of trip requests, we set the step size to 5 minutes instead of the 15 minutes interval so as to deal with the NYC data which is larger than the previous dataset. Likewise, we can make step sizes even smaller to deal with larger demand. One caveat is that the decreasing length of the interval may lead decisions in each batch to become more myopic and deteriorate the solution quality as we utilize restricted information. 

\begin{table}[h]
 \footnotesize
\centering
\caption{Parameter settings for New York City dataset}
\begin{tabular}{|l|l|l|l|l|}
\hline
\textbf{Parameter} & \textbf{Scenario 1}& \textbf{Scenario 2} \\ \hline 
Data & 1\% sampled & 20\% sampled \\ \hline
Fleet size ($M$) & 3, 4, 5, 6 & 30, 40, 50, 60 \\ \hline
Step size & 5 (min)  & 5 (min) \\ \hline
RH factor & 0,1,2,3 & 0,1,2,3 \\ \hline
\end{tabular}%
\label{tab:parameters_NYC}
\end{table}

\subsection{Results of Chattanooga paratransit dataset}\label{section:Chatta}

\begin{figure}[h!]
  \centering
  \includegraphics[scale=0.32]{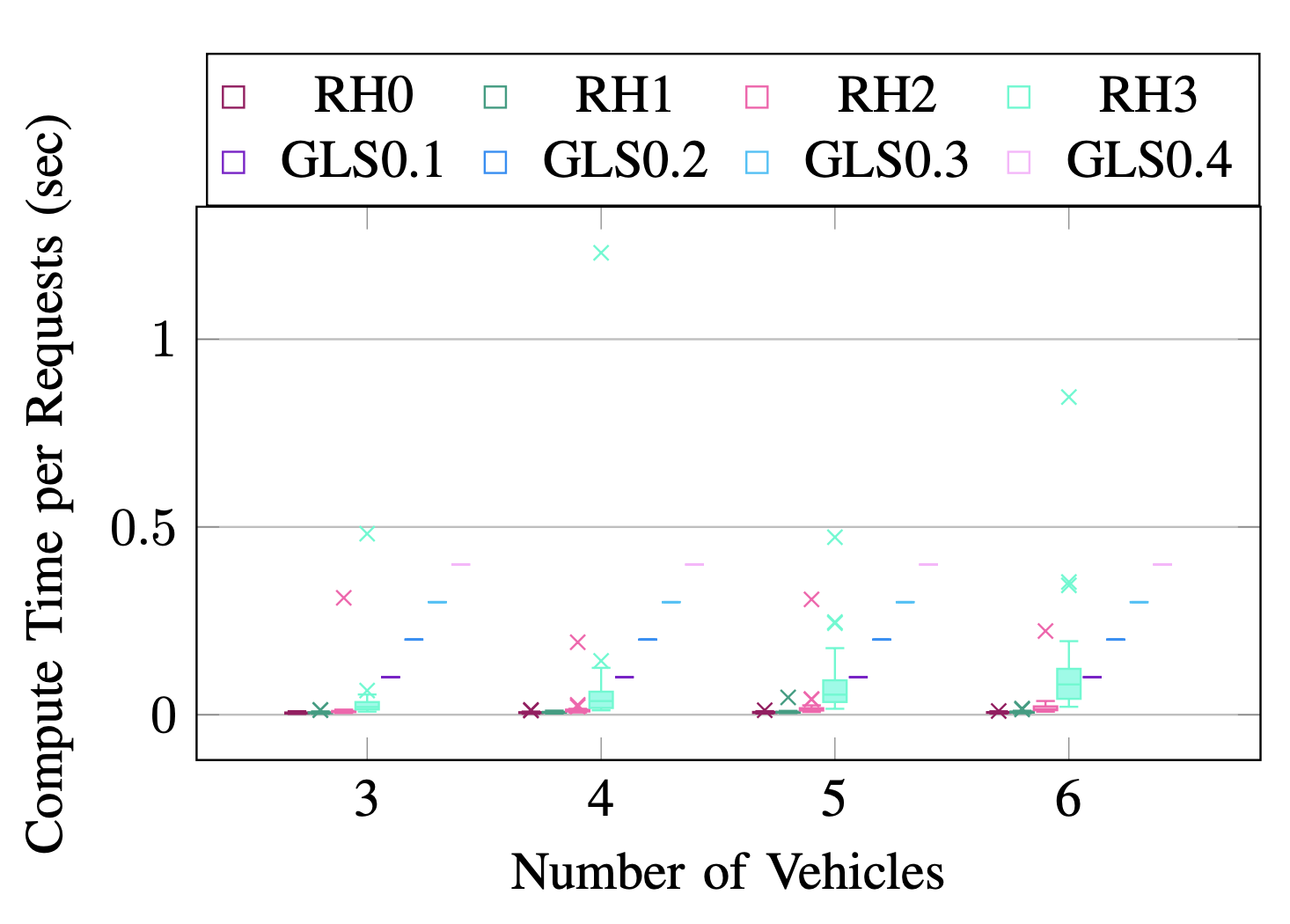}
  \caption{Compute time, Chattanooga paratransit dataset}
  \label{fig:ChattaComputeTime}
\end{figure}

\begin{figure}[h!]
  \centering
  \includegraphics[scale=0.32]{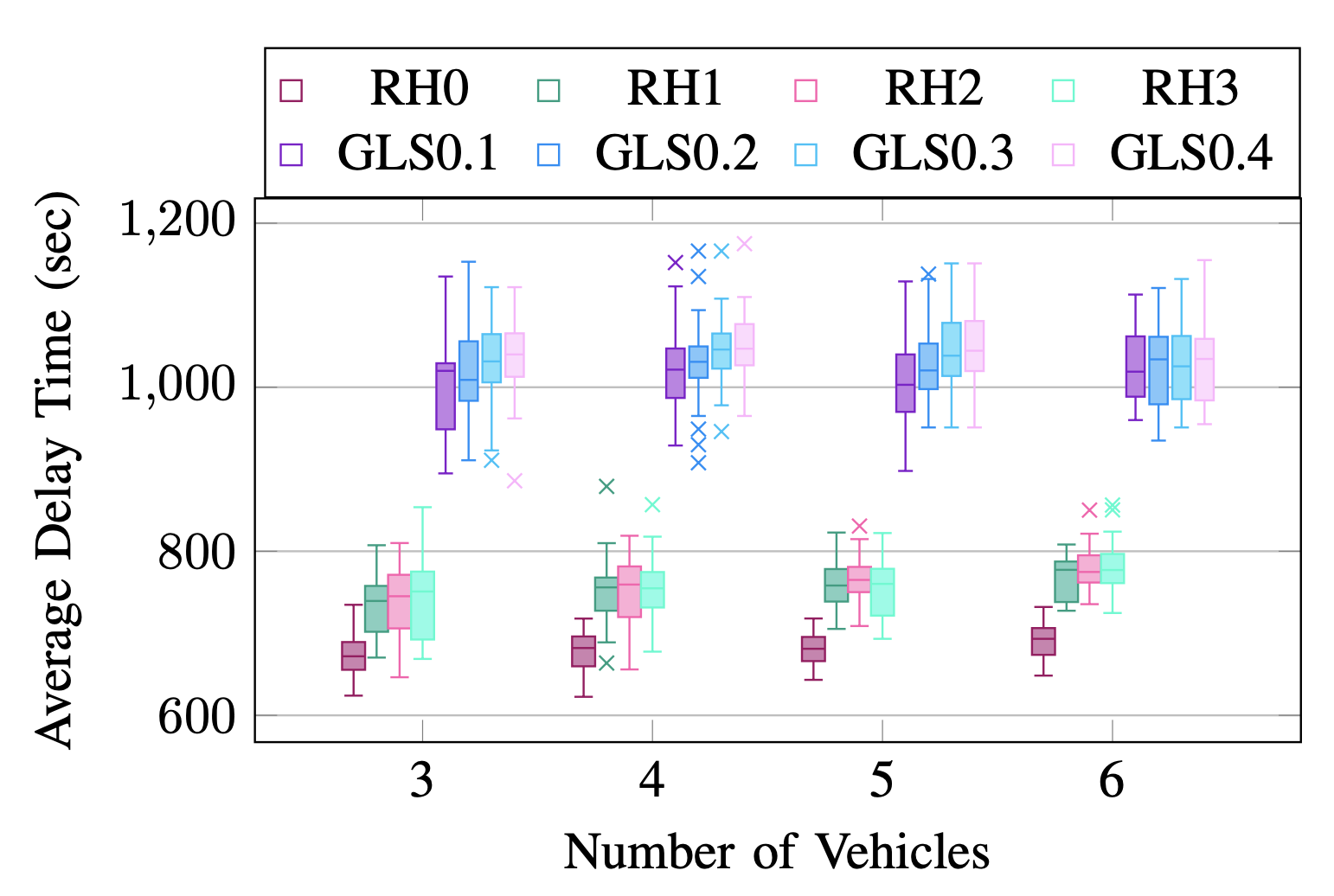}
  \caption{Delay time, Chattanooga paratransit dataset}
  \label{fig:ChattaDelayTime}
\end{figure}

Overall, the trend of the results from the Chattanooga paratransit dataset shows the same as the results from New York City taxi dataset.

\subsection{Pseudo code}\label{section:pseudocode}

\RestyleAlgo{boxruled}

Algorithm \ref{algo:OfflineScheduling} shows the overview of our offline PDPTW algorithm. Starting from the initial simulation time, we iteratively call the functions 
\textbf{WindowProcessing}, \textbf{RTV-ILP}, and \textbf{SimulateVehicle} until the end of the simulation time.

\begin{algorithm}[h!]
 \caption{$\textbf{OfflineScheduling}(\cal{R}, \cal{V})$}
 \label{algo:OfflineScheduling}
 
$t \leftarrow 0$

$\hat{\cal{R}} \leftarrow \emptyset$

\While {$t<t_{max}$}
{
    $ \ActiveRequestSet \leftarrow \ActiveRequestSet \cup \textbf{ WindowProcessing}(t)$
    
    $\Assignment \leftarrow \textbf{RTV-ILP} (\ActiveRequestSet, \VehicleSet)$
    
    ${\cal{P}}_{v \in \VehicleSet,t}, \VehicleRoutes_t \leftarrow \textbf{SimulateVehicle}(\Assignment, t)$
    
    \For {$v \in \VehicleSet$}
    {
    $\ActiveRequestSet \leftarrow \ActiveRequestSet$  $\setminus {\cal{P}}_{v, t}$
    }

    $t \leftarrow t + \Delta t$
}

\KwResult{$\VehicleRoutes$}

\end{algorithm}

Algorithm \ref{algo:WindowProcessing} shows a pseudo code for window processing. Rolling horizon factor \RHFactor is introduced to select an eligible set of requests \NewRequestSet. The procedure for selecting a set of requests to be considered based on their desired pickup time $\textbf{PickupTime}(r_k)$. The window processing makes \NewRequestSet to contain requests within the sliding window size $T_w$ . 

\begin{table*}[h]
\centering
\begin{tabular}{|l|ll|llllll|l|}
\hline
Instance & \multicolumn{2}{c|}{LKH3}                                                    & \multicolumn{6}{c|}{Rolling horizon Framework}                                                                                                                                          & \multicolumn{1}{c|}{Gap (\%)} \\ \hline
         & \multicolumn{1}{c|}{VMT$^a$}           & \multicolumn{1}{c|}{Compute time$^b$} & \multicolumn{2}{c|}{Compute time$^b$}                        & \multicolumn{2}{c|}{Service rate (\%)}                           & \multicolumn{2}{c|}{VMT$^a$}                  & \multicolumn{1}{c|}{}         \\ \hline
         & \multicolumn{1}{l|}{\textbf{}}     & \textbf{}                               & \multicolumn{1}{l|}{$T_w = 5$}  & \multicolumn{1}{l|}{\textbf{$T_w = 10$}}  & \multicolumn{1}{l|}{$T_w = 5$}     & \multicolumn{1}{l|}{\textbf{$T_w = 10$}} & \multicolumn{1}{l|}{$T_w = 5$}  & \textbf{$T_w = 10$}  &                               \\ \hline
lc101    & \multicolumn{1}{l|}{\textbf{997}}  & \textbf{12.05}                          & \multicolumn{1}{l|}{0.42} & \multicolumn{1}{l|}{\textbf{0.44}} & \multicolumn{1}{l|}{100}     & \multicolumn{1}{l|}{\textbf{100}} & \multicolumn{1}{l|}{1095} & \textbf{1127} & 13.00                         \\ \hline
lc105    & \multicolumn{1}{l|}{\textbf{1011}} & \textbf{15.92}                          & \multicolumn{1}{l|}{0.50} & \multicolumn{1}{l|}{\textbf{0.50}} & \multicolumn{1}{l|}{98}      & \multicolumn{1}{l|}{\textbf{100}} & \multicolumn{1}{l|}{1116} & \textbf{1140} & 12.73                         \\ \hline
lc106    & \multicolumn{1}{l|}{\textbf{1032}} & \textbf{22.81}                          & \multicolumn{1}{l|}{0.49} & \multicolumn{1}{l|}{\textbf{0.50}} & \multicolumn{1}{l|}{100}     & \multicolumn{1}{l|}{\textbf{100}} & \multicolumn{1}{l|}{1172} & \textbf{1163} & 12.69                         \\ \hline
lc107    & \multicolumn{1}{l|}{\textbf{1021}} & \textbf{18.51}                          & \multicolumn{1}{l|}{0.49} & \multicolumn{1}{l|}{\textbf{0.60}} & \multicolumn{1}{l|}{98.03} & \multicolumn{1}{l|}{\textbf{100}} & \multicolumn{1}{l|}{1165} & \textbf{1085} & 6.23                          \\ \hline
lc108    & \multicolumn{1}{l|}{\textbf{1030}} & \textbf{18.81}                          & \multicolumn{1}{l|}{0.50} & \multicolumn{1}{l|}{\textbf{0.52}} & \multicolumn{1}{l|}{96.15} & \multicolumn{1}{l|}{\textbf{100}} & \multicolumn{1}{l|}{1209} & \textbf{1120} & 8.74                          \\ \hline
lc201    & \multicolumn{1}{l|}{\textbf{1779}} & \textbf{50.85}                          & \multicolumn{1}{l|}{0.39} & \multicolumn{1}{l|}{\textbf{0.40}} & \multicolumn{1}{l|}{100}     & \multicolumn{1}{l|}{\textbf{100}} & \multicolumn{1}{l|}{1981} & \textbf{2021} & 13.57                         \\ \hline
\end{tabular}
\caption{Comparison of LKH3 and Rolling Horizon Framework ($t_s$ is set to 5 min; $T_w$ is set to 5 and 10 min).}
\footnotesize{$^a$Abbreviation for vehicle miles traveled; $^b$The unit of the compute time is second.}
\label{table:LKH3}
\end{table*}

\begin{algorithm}[h!]
 \caption{$\textbf{WindowProcessing}(t)$}
 \label{algo:WindowProcessing}

\NewRequestSet$ \leftarrow \emptyset$

\If{$t == 0$}
{   
    \For {$r_k \in R$}
    {
        \If{$\textbf{PickupTime}(r_k) \leq \RHFactor \times t_s$}
            {
            $\NewRequestSet  \leftarrow \NewRequestSet \cup \{ r_k \}$
            }
    }
}
\Else{
    \For {$r_k \in R$}
    {
        \If{$\textbf{PickupTime}(r_k) \leq (t + \RHFactor \times t_s)$ $\wedge \textbf{PickupTime}(r_k) > (t + \RHFactor - 1) \times t_s$}
            {
            $\NewRequestSet \leftarrow \NewRequestSet \cup \{ r_k \}$
            }
    }
}
\KwResult{\NewRequestSet}
\end{algorithm}

\subsection{Comparison to LKH3}\label{section:LKH3}

We compare our solver to an extension of Lin-Kernighan-Helsgaun (LKH3). LKH3 is the state-of-the-art solver to solve TSP and its variants. Among 222 PDPTW benchmark instances, only 3 are less than the best-known solution (BKS), 180 are equal to BKS, and 139 are better than BKS. For more details, readers are encouraged to visit \url{http://webhotel4.ruc.dk/~keld/research/LKH-3/}. 

The benchmark instances are artificially generated for PDPTW having varying lengths of time windows and service times, which may not be the case in the dial-a-ride problem. We modified the PDPTW instances to better represent our problem setting. The entire time horizon varies in different instances and has no unit. We find a time window that corresponds to 30 minutes when we consider the entire horizon as 12 hours. Dwell time for pick up and drop off customers is also set to the value corresponding to 5 minutes in 12 hours. In order to have a fair comparison, we selected six adjusted instances (lc 101, 105, 106, 107, 108, and 201) that LKH3 can find routes to serve all customers without violating any demand and time window constraint.

Unlike our solver imposing hard constraints for demands and time windows, LKH3 has soft constraints. LKH3 defines a penalty to measure how much the constraints are violated. The primary objective is to minimize the penalty and the secondary objective is to minimize cost, so if a solution route has a positive penalty, it means the solver could not find such a route that does not violate any constraint. For the instances that LKH3 shows a 100\% service rate with a positive value of penalty, our solver drops some customers because of infeasibility. For those instances, a fair comparison between our solver and LKH3 is hard. Hence, we will focus on the solutions with no penalty.

Table \ref{table:LKH3} compares the performance of our framework to LKH3. The total vehicle miles traveled (VMT) and computation time are used as metrics. The VMT of the rolling horizon framework is higher than the VMT of LKH3 with the optimality gaps varying from 6.23\% to 13.57\%. However, the rolling horizon framework is much faster than LKH3. While LKH3 takes 12.05 sec to 50.85 sec, the rolling horizon framework takes less than 0.6 seconds for all instances. In other words, the rolling horizon framework is around 20 to 80 times faster than the benchmark solver.

}
\fi

\end{document}